\documentclass[10pt,twocolumn,letterpaper]{article}

\usepackage{cvpr}
\usepackage{times}
\usepackage{epsfig}
\usepackage{graphicx}
\usepackage{amsmath}
\usepackage{amssymb}
\usepackage{caption}
\usepackage{comment}
\usepackage{url}
\usepackage[breaklinks=true,bookmarks=false]{hyperref}

\cvprfinalcopy 

\def\cvprPaperID{108} 

\ifcvprfinal\fi

\begin{document}

\title{WiCV at CVPR 2025: The Women in Computer Vision Workshop}

\author{Estefania Talavera$^1$, Deblina Bhattacharjee$^2$, Himangi Mittal$^3$, Mengwei Ren$^4$, Karen Sanchez$^5$,\\
Carla Muntean$^6$, JungEun Kim$^{7,8}$, Mona Jalal$^9$. \\\\ 
$^1$University of Twente, $^2$University of Bath, $^3$Carnegie Mellon University, $^4$Adobe, $^5$KAUST, \\
$^6$Microsoft, $^7$KAIST, $^8$General Robotics, $^9$Toyota Material Handling North America.\\ 
\tt\small wicvcvpr2025-organizers@googlegroups.com
}
\maketitle
\begin{abstract}
\thispagestyle{empty}
The Women in Computer Vision Workshop (WiCV@CVPR 2025) was held in conjunction with the IEEE/CVF Conference on Computer Vision and Pattern Recognition (CVPR 2025) in Nashville, Tennessee, United States. This report presents an overview of the workshop program, participation statistics, mentorship outcomes, and historical trends from previous WiCV editions. The goal is to document the impact and evolution of WiCV as a reference for future editions and for other initiatives aimed at advancing diversity, equity, and inclusion within the AI and computer vision communities.
WiCV@CVPR 2025 marked the 16th edition of this long-standing event dedicated to increasing the visibility, inclusion, and professional growth of women and underrepresented minorities in the computer vision community. This year's workshop featured 14 accepted papers in the CVPR Workshop Proceedings out of 32 full-paper submissions. Five of these were selected for oral presentations, while all 14 were also presented as posters, along with 36 extended abstract posters accepted from 62 short-paper submissions, which are not included in the proceedings.
The mentoring program matched 80 mentees with 37 mentors from both academia and industry. The 2025 edition attracted over 100 onsite participants, fostering rich technical and networking interactions across all career stages. Supported by 10 sponsors and approximately \$44,000 USD in travel grants and diversity awards, WiCV continued its mission to empower emerging researchers and amplify diverse voices in computer vision.
\end{abstract}

\section{Introduction}
Despite remarkable progress in various areas of computer vision research in recent years, the field still grapples with a persistent lack of diversity and inclusion. Although the field of computer vision rapidly expands, female researchers remain underrepresented in the area, comprising only a small number of professionals in both academia and industry. As a result, many women in computer vision experience isolation in environments that remain unbalanced.

The WiCV workshop is a gathering designed for all individuals, regardless of gender, engaged in computer vision research. It aims to appeal to researchers at all levels, including established researchers in both industry and academia (e.g., faculty or postdocs), graduate students pursuing a Master's or PhD, as well as undergraduates interested in research.  The overarching goal is to enhance the visibility and recognition of female computer vision researchers across these diverse career stages, reaching women from various backgrounds in educational and industrial settings worldwide.

There are three key objectives of the WiCV workshop:

\subsection{Networking and Mentoring} The first objective is to expand the WiCV network and facilitate interactions between members of this network. This includes female students learning from experienced professionals who share career advice and experiences. A mentoring banquet held alongside the workshop provides a casual environment for junior and senior women in computer vision to meet, exchange ideas, and form mentoring or research relationships.

\subsection{Raising Visibility}

The workshop's second objective is to elevate the visibility of women in computer vision, both at junior and senior levels. Senior researchers are invited to give high-quality keynote talks on their research, while junior researchers are encouraged to submit their recent or ongoing work, with many of these being selected for oral or poster presentation through a rigorous peer review process. This empowers junior female researchers to gain experience presenting their work in a professional yet supportive setting. The workshop aims for diversity not only in research topics but also in the backgrounds of presenters. Additionally, a panel discussion provides a platform for female colleagues to address topics of inclusion and diversity.

\subsection{Supporting Junior Researchers}

Finally, the third objective is to offer junior female researchers the opportunity to attend a major computer vision conference that might otherwise be financially inaccessible. This is made possible through travel grants awarded to junior researchers who present their work during the workshop's poster session. These grants not only enable participation in the WiCV workshop but also provide access to the broader CVPR conference.

\section{Workshop Program Overview}
\label{program}
The Women in Computer Vision Workshop (WiCV@CVPR 2025) offered a half-day hybrid program combining four keynote talks, five oral presentations, 41 posters, a panel discussion, and a mentorship program. The event aimed to foster both technical exchange and community building among researchers at all career stages.

\subsection{Invited Talks}

Following the tradition of previous editions, the program was designed to highlight a wide range of technical topics and career experiences, fostering inclusion, visibility, and mentorship across all professional levels.

WiCV@CVPR 2025 featured four invited speakers from academia and industry who shared insights on research, leadership, and diversity in computer vision (see Table \ref{keynote_speakers}). Each talk was followed by Q$\&$A sessions that encouraged dialogue across disciplines and experience levels. These talks spanned emerging topics such as foundation models, trustworthy AI, and multimodal perception systems, ensuring an interdisciplinary dialogue.

\subsection{Oral and Poster Sessions}

The oral and poster sessions formed the technical core of the WiCV@CVPR 2025 program, showcasing a diverse range of research topics that reflect current trends and challenges in computer vision. The oral presentations highlighted high-impact work spanning generative modeling, adversarial robustness, medical image analysis, and vision–language reasoning. These talks provided visibility to early-career researchers and fostered in-depth technical discussions among participants.

The poster sessions complemented the oral program by featuring a broad spectrum of ongoing research and preliminary studies. Presenters shared innovative approaches in areas such as transformer-based architectures, 3D reconstruction, medical and biological imaging, and privacy-preserving AI. The interactive poster format encouraged direct exchange between junior and senior researchers, offering an inclusive environment for feedback, collaboration, and mentorship.

Table~\ref{table1} lists the oral presenters and their paper titles, illustrating the high quality and diversity of topics represented at the workshop.

\begin{table}[]
\caption{List of oral presenters and their corresponding paper titles presented at WiCV@CVPR 2025.}
\begin{tabular}{ll}
\hline
\multicolumn{1}{c}{\textbf{Presenter}} & \multicolumn{1}{c}{\textbf{Paper Title}}  \\ \hline
Mika Feng   & \begin{tabular}[c]{@{}l@{}}Leveraging Intermediate Features of Vision\\ Transformer for Face Anti-Spoofing\end{tabular}                                                 \\ 
\begin{tabular}[c]{@{}l@{}}Pooja\\ Kumari\end{tabular}     & \begin{tabular}[c]{@{}l@{}}Document Image Rectification using Stable\\ Diffusion Transformer\end{tabular} \\ 
\begin{tabular}[c]{@{}l@{}}Amanda N.\\ Nikho\end{tabular}  & \begin{tabular}[c]{@{}l@{}}Knowledge Distillation Approach for SOS\\ Fusion Staging: Towards Fully Automated\\ Skeletal Maturity Assessment\end{tabular} \\ 
\begin{tabular}[c]{@{}l@{}}Suruchi\\ Kumari\end{tabular}   & \begin{tabular}[c]{@{}l@{}}Leveraging Fixed and Dynamic Pseudo-\\ Labels in Cross-Supervision Framework\\ for Semi-Supervised Medical Image\\ Segmentation\end{tabular} \\ 
\begin{tabular}[c]{@{}l@{}}Nurjahan\\ Sultana\end{tabular} & \begin{tabular}[c]{@{}l@{}}Domain Adaptation for Skin Lesion:\\ Evaluating Real-World Generalisation\end{tabular} \\ \hline
\end{tabular}
\label{table1}
\end{table}

The oral sessions covered topics ranging from adversarial robustness and medical image analysis to generative models and cross-domain adaptation.

\begin{table}[]
\caption{Summary of WiCV@CVPR 2025 submissions, acceptance rates, and presentation modes for full papers and extended abstracts included in the program proceedings.}
\scalebox{0.75}{
\begin{tabular}{lllll}
\hline
\multicolumn{1}{c}{\textbf{Type}}                         & \multicolumn{1}{c}{\textbf{Submissions}} & \multicolumn{1}{c}{\textbf{Accepted}} & \multicolumn{1}{c}{\textbf{\begin{tabular}[c]{@{}c@{}}Acceptance\\ Rate (\%)\end{tabular}}} & \multicolumn{1}{c}{\textbf{\begin{tabular}[c]{@{}c@{}}Presentation\\ Mode\end{tabular}}} \\ \hline
\begin{tabular}[c]{@{}l@{}}Full Papers- In\\ Proceedings\end{tabular} & 32        & 14  & 43.8         & \begin{tabular}[c]{@{}l@{}}5 oral,\\ 14 posters\end{tabular}  \\ 
\begin{tabular}[c]{@{}l@{}}Extended\\ Abstracts\end{tabular} & 62  & 36 & 58.06 & Posters\\ \hline
\end{tabular}}
\label{table2}
\end{table}

\subsection{Panel Discussion}

A highly engaging panel brought together four leading researchers, Roni Sengupta, Olga Russakovsky, Xiaoying Jin, and Petia Radeva, to discuss Building Inclusive Communities in Computer Vision.
The conversation emphasized strategies for mentorship, equitable research visibility, and structural inclusion in AI communities.

\subsection{Mentorship and Networking Sessions}

Following the formal sessions, WiCV hosted an in-person mentoring dinner featuring two invited talks by Petia Radeva (University of Barcelona) and Neda Hantehzadeh (Director of Data Science, CCC Intelligent Solutions).
The mentoring program included 80 mentees and 37 mentors from academia and industry: Gul Varol, Houda Chini, Bijie Qiu, Melissa Hall, Jiekai Ma, Kaustubha, Nikhila Ravi, Negar Rostamzadeh, Deepti Ghadiyaram, Adriana Romero Soriano, Ana-Maria Marcu, Roni Sengupta, Petia Radeva, Heba Sailem, Akshata Kishore Moharir, Ishmeet Kaur, Mingfei Yan, Fazilet Gokbudak, Either Oduntan, Iqra Ali, Medha Sawhney, Jimoh Abdulganiyu, Aura Arefeh Yavary, Nahid Alam, Judy Hoffman, Boyi Li, Arpita Vats, Ankita Shukla, Vaishnavi Yeruva, Carlos Hinojosa, Kohinoor Roy, Subarna Tripathi, Kate Saenko, Anuradha Kar, Efstathia Soufleri, Karen Sanchez.
This mentoring dinner included the Sponsors Exhibition (in person).

\subsection{Hybrid Setting}
The 2025 edition adopted a hybrid format, with all oral sessions, invited talks, and the panel streamed live. The onsite events took place at the Music City Center in Nashville, Tennessee, while online participants joined through Zoom.

\begin{table*}[ht!]
\caption{List of keynote speakers at WiCV@CVPR 2025, including their affiliations and talk titles.}
\begin{tabular}{lll}
\hline
\multicolumn{1}{c}{\textbf{Speaker}} & \multicolumn{1}{c}{\textbf{Affiliation}} & \multicolumn{1}{c}{\textbf{Talk Title / Topic}}           \\ \hline
Dr. Xiaoying Jin                       & \begin{tabular}[c]{@{}l@{}}Senior Manager, Computer Vision and\\ AI/ML Engineering, HERE Technologies\end{tabular}                   & \begin{tabular}[c]{@{}l@{}}Revolutionizing Mapmaking with Perception,\\ Language, and Generative AI\end{tabular}        \\ 
Dr. Roni Sengupta                      & \begin{tabular}[c]{@{}l@{}}Assistant Professor of Computer Science at the\\ University of North Carolina at Chapel Hill\end{tabular} & \begin{tabular}[c]{@{}l@{}}Learning to See in the Light: Foundation Models\\ for Lighting-Aware Perception\end{tabular} \\ 
MSc Nikhila Ravi & Research Lead at Meta AI                          & \begin{tabular}[c]{@{}l@{}}Building Towards General-Purpose Perception\\ Systems\end{tabular}                           \\ 
Dr. Olga Russakovsky                   & \begin{tabular}[c]{@{}l@{}}Associate Professor, Computer Science\\ Department, Princeton University\end{tabular}                     & Trustworthy (and trusted) computer vision       \\ \hline
\end{tabular}
\label{keynote_speakers}
\end{table*}

\section{Participation and Review Statistics}

\subsection{Submissions and Acceptance}

The first edition of the Women in Computer Vision (WiCV) workshop was held in conjunction with CVPR 2015. Over the years, both the participation rate and the quality of submissions to WiCV have steadily increased. Following the examples from previous editions \cite{asra2023wicv, antensteiner2022wicv, doughty2021wicv, Amerini19, Akata18, Demir18}, we have continued to curate top-quality submissions into our workshop proceedings. By providing oral and poster presenters the opportunity to publish their work in the conference proceedings, we aim to further boost the visibility of female researchers.

This year, we received 94 high-quality submissions covering a wide range of topics and institutions, a slight increase compared to WiCV@CVPR24. The most popular topics included deep learning architectures and vision transformers for image analysis, segmentation, and restoration; medical and biological imaging (e.g., glaucoma detection, fetal ultrasound, glioma segmentation, and low-dose X-ray enhancement); 3D vision and diffusion-based modeling (e.g., 3D reconstruction, Gaussian splatting, tri-plane deformation); vision–language models for multimodal reasoning and self-correction; biometrics and face analysis (e.g., morphing attack detection, kinship verification, anti-spoofing); adversarial learning and privacy-preserving AI addressing robustness and data reconstruction; and remote sensing and ecological monitoring.

\begin{figure}[h]
\centering
\includegraphics[width=1\linewidth]{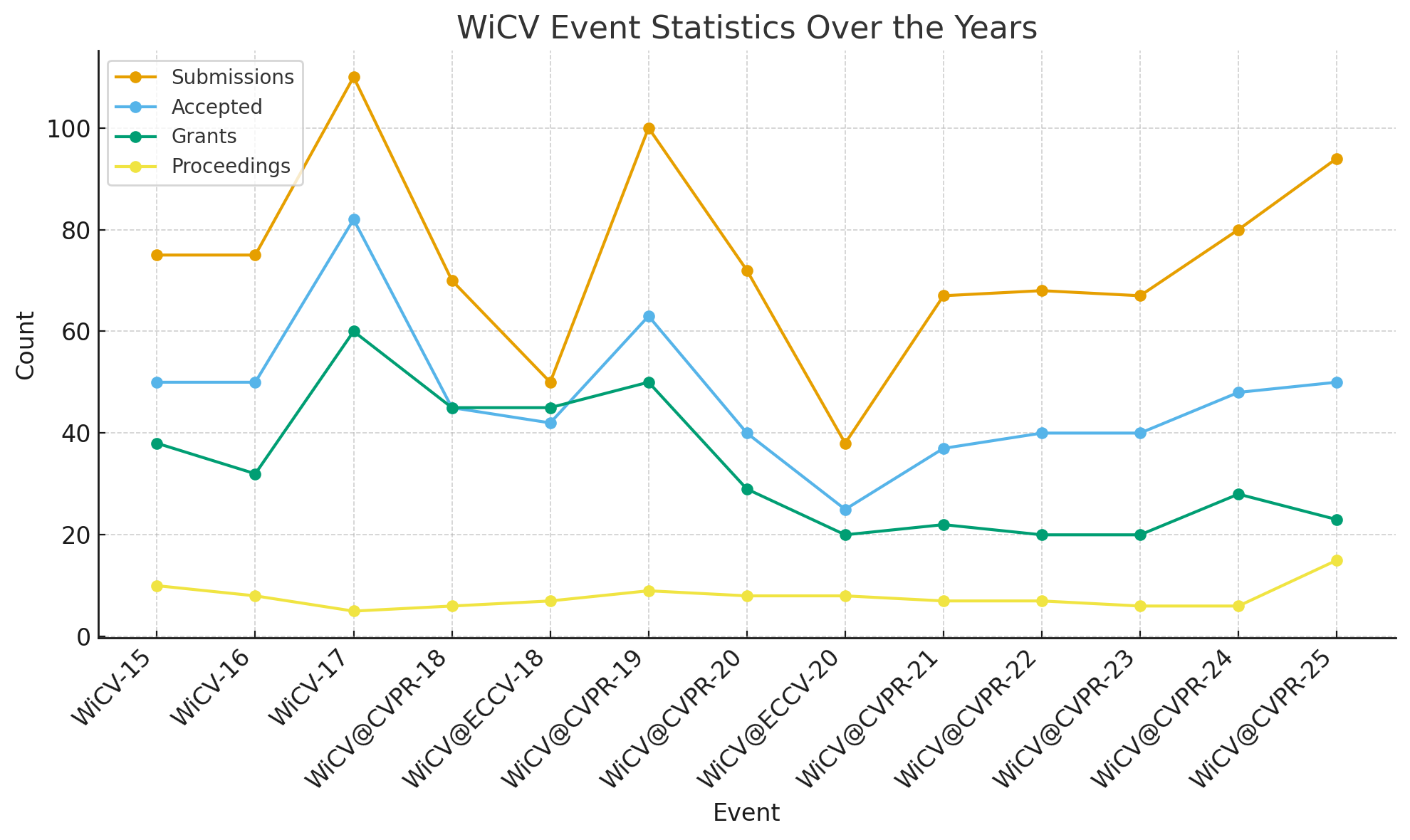}
\captionof{figure}{Number of submissions across \textbf{WiCV} editions, illustrating the continued growth of participation.}
\label{fig:sub}
\end{figure}

The WiCV@CVPR 2025 workshop received a total of 94 submissions across two categories (Table~\ref{table2}): 32 full-paper submissions and 62 short-paper submissions under the extended abstract modality. Of these, 81 underwent the review process. Fourteen full papers were accepted for inclusion in the CVPR Workshop Proceedings, with five selected for oral presentations and all 14 also showcased as posters. Additionally, 36 extended abstracts were accepted for poster presentations, though these were not included in the proceedings. Table~\ref{table1} lists the top five-ranked papers, their presenters, and titles, while Figure~\ref{fig:sub} presents a comparative summary with previous years.

\subsection{Review and Evaluation Process}

All submissions were reviewed by an interdisciplinary program committee composed of 41 expert reviewers. Each submission received multiple detailed evaluations, ensuring fairness and depth in the review process. Reviewers provided constructive feedback, which was shared with authors to improve their work and to foster learning and exchange within the community.

The diligent effort of the program committee ensured the high quality of accepted papers, continuing WiCV’s tradition of promoting rigorous and impactful research across a broad range of computer vision domains.

\subsection{Participation and Mentorship}

The 2025 WiCV workshop was held as a half-day in-person event with hybrid options, hosted at the Music City Center in downtown Nashville, Tennessee, United States. The virtual component was facilitated via Zoom, ensuring accessibility and global participation.

Senior and junior researchers were invited to present their work through oral and poster sessions, as detailed in Section~\ref{program}. The organizing team brought together members from diverse academic and industrial backgrounds across different time zones, enriching the workshop’s organization with a wide range of expertise. Their research interests spanned deep-learning architectures for image and video understanding, multimodal vision–language systems, medical and biological imaging, and privacy-preserving AI.

During the mentoring session, 80 mentees received in-person guidance from 37 mentors. To accommodate authors who could not attend in person, a two-hour online mentoring session was organized via Zoom, attended by 16 participants. These activities exemplified WiCV’s continued commitment to fostering mentorship, inclusivity, and community growth.

\subsection{Sponsorship \& Travel Grants}

Following the tradition established in previous editions \cite{Akata18,Amerini19,Demir18,doughty2021wicv, goel2022wicv,aslam2024wicv}, WiCV@CVPR 2025 provided travel and participation grants to support authors of accepted submissions. These grants were designed to alleviate financial barriers and enable attendance at the workshop. Funding covered a range of expenses, depending on the needs of each participant, including conference registration fees, round-trip flights, and accommodation for up to two days.
WiCV@CVPR 2025 secured $\$44,000$ USD in sponsorship. A total of 10 sponsors contributed to the event. The San Francisco Study Center served as the fiscal sponsor, assisting with the management of sponsorships and travel awards. Through these sponsorship efforts, WiCV continues to ensure equitable access and empower the participation of underrepresented groups in computer vision research.

\section{Conclusions}

WiCV@CVPR 2025 once again demonstrated the importance of community-driven efforts in fostering diversity, inclusion, and visibility within the computer vision field. The workshop successfully brought together researchers from academia and industry, spanning all career stages, to share their work and experiences. Through an engaging program of oral and poster sessions, mentoring activities, and panel discussions, WiCV continued to provide a supportive environment that amplifies underrepresented voices and encourages professional growth.

Building on the success of previous editions, WiCV has strengthened its impact through expanded participation, interdisciplinary discussions, and sustained mentoring initiatives. With a growing number of high-quality submissions and dedicated sponsorship support, the workshop continues to grow as a global platform for equitable participation in computer vision. Looking ahead, the WiCV community remains committed to advancing diversity, equity, and inclusion, and to inspiring the next generation of researchers who will shape the future of the field.

\section{Acknowledgments}
We express our sincere gratitude to our sponsors, including our Platinum sponsors: Apple and Meta; as well as our Gold Sponsor: King Abdullah University of Science and Technology KAUST; Silver Sponsor: Wayve; and Bronze sponsors: RBC Borealis, Google DeepMind, Meshcapade, Zoox, NVIDIA, and Tencent.
Our appreciation also extends to the San Francisco Study Center, our fiscal sponsor, for their invaluable assistance in managing sponsorships and travel awards. We are thankful for the support and knowledge sharing from organizers of previous WiCV workshops, without whom this WiCV event would not have been possible. Finally, we extend our heartfelt thanks to the dedicated program committee, authors, reviewers, submitters, and all participants for their valuable contributions to the WiCV network community.

\section{Contact}
\noindent \textbf{Website}: \url{https://sites.google.com/view/wicv-cvpr-2025}\\
\textbf{E-mail}: wicvcvpr2025-organizers@googlegroups.com\\
\textbf{X}: \url{https://x.com/wicvworkshop}\\
\textbf{Google group}: women-in-computer-vision@googlegroups.com\\

{\small
\bibliographystyle{ieee}
\bibliography{egbib}
}

\end{document}